%% file: main.tex
\pdfoutput=1

\documentclass[11pt,a4paper]{article}
\usepackage[nohyperref]{latex/emnlp2018}
\usepackage{times}

\usepackage{url}

\aclfinalcopy % Uncomment this line for the final submission

\usepackage{flushend}

\usepackage[T1]{fontenc}
\usepackage{tgtermes}
\usepackage{tgheros}             % Helvetica (sans-serif)
\usepackage{tgcursor}            % Courier   (typewriter)

\usepackage[protrusion=true,expansion=true]{microtype}

\usepackage{latexsym}
\usepackage{graphicx}

\usepackage{array}
\usepackage{booktabs}
\usepackage{tabularx}
\usepackage{multirow}

\usepackage{siunitx}
\usepackage[margin=0pt,singlelinecheck=off]{caption}
\usepackage{amsmath, amssymb}
\usepackage{enumitem}

\usepackage{algorithm}
\usepackage{algpseudocode}
\usepackage{xspace}

\input{definitions.tex}

\title{Compact Personalized Models for Neural Machine Translation}

\author{Joern Wuebker, Patrick Simianer, John DeNero\\
  Lilt, Inc. \\
  {\tt first\_name@lilt.com} \\}

\begin{document}

\maketitle
\begin{abstract}
\input{0-abstract}
\end{abstract}

\input{1-intro}

\input{2-related}

\input{3-transformer}

\input{4-adapt}

\input{experiments}

\input{9-conclusion}

% \section*{Acknowledgements}

%%%%%%%%%%%%%%%%%%%%%%%%%%%
%\newpage
\bibliography{translation}

\end{document}

%% file: definitions.tex
\usepackage{xcolor}

\definecolor{orange}{rgb}{1.0, 0.6, 0.0}
\definecolor{yellow}{rgb}{1.0, 1.0, 0.0}

\definecolor{grey}{rgb}{0.6, 0.6, 0.6}
\definecolor{lightgrey}{rgb}{0.9, 0.9, 0.9}

\def\blue#1    {\textcolor{blue} {#1} }
\def\green#1   {\textcolor{green} {#1} }
\def\magenta#1 {\textcolor{magenta} {#1} }

\newcommand\BLEU{\textsc{Bleu}\xspace}

\newcommand\emptyword\epsilon

\newcommand\ngrams{$n$-grams\xspace}

\newcommand\mgrams\ngrams

\newcommand{\sourceword}{\ensuremath{f}\xspace}
\newcommand{\targetword}{\ensuremath{e}\xspace}
\newcommand\f\sourceword
\newcommand\e\targetword
\newcommand{\segphrase}{\ensuremath{s}\xspace}
\newcommand{\alignmentword}{\ensuremath{a}\xspace}
\newcommand{\sourcemax}{\ensuremath{J}\xspace}
\newcommand{\targetmax}{\ensuremath{I}\xspace}
\newcommand{\segmax}{\ensuremath{K}\xspace}
\newcommand{\sourceindex}{\ensuremath{j}\xspace}
\newcommand{\targetindex}{\ensuremath{i}\xspace}
\newcommand{\segindex}{\ensuremath{k}\xspace}

\newcommand\invalignmentword{\ensuremath{B}\xspace}

\renewcommand\j\sourceindex
\renewcommand\i\targetindex
\renewcommand\k\segindex
\newcommand\J\sourcemax
\newcommand\I\targetmax
\newcommand\K\segmax

\newcommand{\sourcephrase}{\ensuremath{\tilde{\sourceword}}\xspace}
\newcommand{\targetphrase}{\ensuremath{\tilde{\targetword}}\xspace}
\newcommand{\sourcesentence}{\ensuremath{\f_1^\J}\xspace}
\newcommand{\targetsentence}{\ensuremath{\e_1^\I}\xspace}
\newcommand{\segsentence}{\ensuremath{\s_1^\K}\xspace}

\newcommand{\alignmentsentence}{\ensuremath{\a_1^{\J}}\xspace}
\newcommand{\invalignmentsentence}{{\ensuremath{\b_0^{\I}}}\xspace}

\newcommand\s\segphrase
\renewcommand\a\alignmentword
\renewcommand\b\invalignmentword
\newcommand\fs\sourcesentence
\newcommand\fp\sourcephrase
\newcommand\es\targetsentence
\newcommand\ep\targetphrase
\newcommand\sse\segsentence
\newcommand\as\alignmentsentence
\newcommand\bs\invalignmentsentence

%% file: 0-abstract.tex
We propose and compare methods for gradient-based domain adaptation of self-attentive neural machine translation models.
We demonstrate that a large proportion of model parameters can be frozen during adaptation with minimal or no reduction in translation quality by
encouraging structured sparsity in the set of offset tensors during learning via group lasso regularization.
We evaluate this technique for both batch and incremental adaptation across multiple data sets and language pairs.
Our system architecture---combining a state-of-the-art self-attentive model with compact domain adaptation---provides high quality personalized machine translation that is both space and time efficient.

%% file: 1-intro.tex
\section{Introduction}

Professional translators typically translate a collection of related documents drawn from a domain for which they have a set of previously translated examples.
Domain adaptation is critical to providing high quality suggestions for interactive machine translation and post-editing interfaces.
When many translators use the same shared service, the system must train and apply a \emph{personalized} adapted model for each user.
We describe a system architecture and training method that achieve high space efficiency, time efficiency, and translation performance by encouraging structured sparsity in the set of offset tensors stored for each user.

Effective model personalization requires both \emph{batch} adaptation to an in-domain training set, as well as \emph{incremental} adaptation to the test set.
Batch adaptation is applied when a user uploads relevant translated documents before starting to work.
Incremental adaptation is applied when a user provides a correct translation of each segment just after receiving machine translation suggestions, and the system is able to train on that correction before generating suggestions for the next segment. This is referred to as \emph{a posteriori} adaptation by \newcite{turchi2017continuous}.
Our experiments compare both types of adaptation.
There are cases for which incremental adaptation achieves better performance using fewer examples, as examples drawn directly from the test set are often highly relevant to subsequent parts of that test set.
There are also cases for which the gains from both types of domain adaptation are additive.

The time required to translate and to adapt both must be minimal in a personalized translation service.
Interactive translation requires suggestions to be generated at typing speed, and incremental adaptation must occur within a few hundred milliseconds to keep up with a translator's typical workflow.
The service can be expected to store models for a large number of users and dynamically load and adapt models for many active users concurrently.
Therefore, minimizing the number of parameters stored for each user's personalized model is important both for reducing storage requirements and latency.
We achieve space and time efficiency by representing each user's model as an offset from the unadapted baseline parameters and encouraging most offset tensors to be zero during adaptation.

We show that group lasso regularization can be applied to a self-attentive Transformer model to freeze up to 75\% of the parameters with minimal or no loss of adapted translation quality across experiments on four English$\to$German data sets. We confirm these findings for six additional language pairs.

%% file: 2-related.tex
\section{\label{sec:related}Related Work}

There is extensive work on incremental adaptation from human post edits or simulated post edits, both for statistical machine translation \cite{green13:postEditing,denkowski-dyer-lavie:2014:EACL,denkowski-EtAl:2014:HaCaT,wuebker15:incrementalAdaptation} and neural machine translation \cite{DBLP:journals/corr/PerisCC17,turchi2017continuous,DBLP:journals/corr/abs-1712-04853}.
Both \newcite{turchi2017continuous} and \newcite{DBLP:journals/corr/abs-1712-04853} apply 
vanilla fine-tuning algorithms. In addition to fine-tuning towards user corrections, the former applies \emph{a priori} adaptation to retrieved data that is similar to the incoming source sentences. \newcite{DBLP:journals/corr/PerisCC17} propose a variant of fine-tuning with passive-aggressive learning algorithms. In contrast to these papers, where all model parameters are possibly altered during training, this work focuses on space efficiency of the adapted models.

Regularization methods that promote or enforce sparsity have been previously used in the context of sparse feature models for SMT: \newcite{duh10multitask} presented an application of multi-task learning via $\ell_1/\ell_2$ regularization for feature selection in an $N$-best reranking task.
A similar approach, employing $\ell_1/\ell_2$ regularization for feature selection and multi-task learning,  was developed by \newcite{simianer2012joint} and \newcite{simianer-riezler:2013:WMT} for tuning of SMT systems.
Both works report improvements from regularization.

Techniques for enforcing sparse models using $\ell_1$ regularization during stochastic gradient descent optimization were previously developed for linear models \cite{tsuruoka2009stochastic}.

An extremely space efficient method for personalized model adaptation is presented by \newcite{michel:2018:extremeAdaptation}. Here, adaptation is performed solely on the output vocabulary bias vector. Another notable approach for creating compact models is student-teacher-training or knowledge distillation 
\cite{kim-rush:2016:EMNLP2016}. To the best of our knowledge, this has not been applied in a domain adaptation setting.

%% file: 3-transformer.tex
\section{\label{sec:transformer}Self-Attentive Translation Model}

The neural machine translation systems used in this work are based on the Transformer model introduced by \newcite{DBLP:conf/nips/VaswaniSPUJGKP17}, which uses self-attention rather than recurrent or convolutional layers to aggregate information across words.
In addition to its superior performance, its main practical advantage over recurrent models is faster training.

The Transformer follows the encoder-decoder paradigm.
Source word vectors $x_1, \dots, x_m$ are chosen from an embedding matrix $X_e$.
A series of stacked encoder layers generate intermediate representations $z_1, \dots, z_m$.
Each layer of the encoder consists of two sub-layers: a multi-head \emph{self-attention} layer that uses scaled dot-product attention over all source positions, followed by a feed-forward \emph{filter} layer.
Layer normalization \cite{ba2016layer}, dropout \cite{srivastava2014dropout}, and residual connections \cite{he2016deep} are applied to each sub-layer.

A series of stacked decoder layers produces a sequence of target word vectors $y_1, \dots, y_n$.
Each decoder layer has three sub-layers: self-attention, encoder-attention, and a filter.
For target position $j$, the \emph{self-attention} layer can attend to any previous target position $j' \in [1, j]$, with target words offset by one so that representations at $j$ can observe word $j\!-\!1$, but not word $j$.
The \emph{encoder-attention} layer can attend to the final encoder state $z_i$ for any source position $i \in [1,m]$.
Observed target word vectors are chosen from an embedding matrix $Y_e$, and target word $j$ is predicted from $y_j$ via a soft-max layer parameterized by an output projection matrix $Y_o$.

The encoders in this work have six layers that have a \emph{self-attention} sub-layer size of 256 and a \emph{filter} sub-layer size of 512. Each filter performs two linear transformations and a ReLU activation:
$$f(x) = \max(0, xW_1 + b_1)W_2 + b_2 .$$

The decoders in this work have three layers, and all sub-layer sizes are 256. The decoder sub-layers are simplified versions of those described in \newcite{DBLP:conf/nips/VaswaniSPUJGKP17}: The \emph{filter} sub-layers perform only a single linear transformation, and layer normalization is only applied once per decoder layer after the \emph{filter} sub-layer.

Unlike in \newcite{DBLP:conf/nips/VaswaniSPUJGKP17}, none of $X_e$, $Y_e$, or $Y_o$ share parameters in our TensorFlow\footnote{\url{https://www.tensorflow.org/}} implementation.
Baseline models are optimized with Adam \cite{kingma2015adam}.

%% file: 4-adapt.tex
\section{\label{sec:adapt}Compact Adaptation}

\subsection{Fine Tuning}

Personalized machine translation requires batch adaptation to a domain-relevant bitext (such as a user provided translation memory) as well as incremental adaptation to the test set.
We apply fine-tuning, which involves continuing to train model parameters with a gradient-based method on domain-relevant data, as a simple and effective method for neural translation domain adaptation \cite{Luong-Manning:iwslt15}. The fine-tuned model without regularization and clipping is denoted as the \emph{Full Model}.
Confirming previous work, we found that stochastic gradient descent (SGD) is the most effective optimizer for fine tuning \cite{turchi2017continuous}.
In our experiments, batch adaptation uses a batch size of 7000 words for 10 Epochs and a fixed learning rate of 0.1, dropout of 0.1, and label smoothing with $\epsilon_{ls}=0.1$ \cite{szegedy2016rethinking}.

Incremental adaptation uses a batch size of one and a learning rate of 0.01. To ensure a strong adaptation effect within a single document, we set dropout and label smoothing to zero and perform up to three SGD updates on each segment. After each update, we measure the model perplexity on the current training example and continue with another update if the perplexity is still above 1.5. 

\begin{table*}[!tbp]
\begin{center}
\begin{tabular}{l|c|cc|cc|cc|cc}
\toprule
  &  & \multicolumn{2}{c|}{\bf User1 } & \multicolumn{2}{c|}{\bf User2 } & \multicolumn{2}{c|}{\bf Autodesk} & \multicolumn{2}{c}{\bf IWSLT}\\
&\small \# Param. & \small Batch & \small Incr. & \small  Batch & \small Incr. & \small  Batch & \small Incr. & \small  Batch & \small Incr. \\
\midrule
Baseline 		        & 36.2M& \multicolumn{2}{c|}{35.7} & \multicolumn{2}{c|}{32.7} & \multicolumn{2}{c|}{40.3} & \multicolumn{2}{c}{25.9}\\
\midrule
Full Model	      & 25.8M & 47.5 & 48.2 & 44.2 & 34.8 & 47.7 & 46.6 & 27.5 & 26.3\\
\midrule
Outer Layers      & 2.2M  & 45.0 & 47.9 & 36.4 & 33.1 & 45.5 & 44.7 & 27.3  & 26.1\\
Inner Layers          & 2.7M & 45.4 & 47.2 & 36.7 & 33.6 & 45.5 & 43.9 &27.8 &26.5\\
Encoder Embed.       & 5.0M & 41.7	& 41.8 & 33.6 & 32.9 & 42.5 & 41.6 & 27.4 & 26.4\\
Decoder Embed.        & 5.5M & 36.6	& 37.6 & 33.0 & 33.0 & 40.8 & 40.5 & 26.2 & 25.9\\
Output Proj.       & 10.3M & 44.2 & 46.0 & 38.1 & 34.5 & 45.2 & 42.9 & 27.1 & 26.5 \\
Sparse Output Proj. (*) & 5.5M & 43.5 & 46.7  & 39.7 & 34.7  & 45.5 & 43.3 & 27.1 & 26.7 \\
\midrule
   (*) + Fixed & 6.9M & 46.4 & 47.8 & 42.3 & 30.9  & 47.6 & 43.7 & 27.3 & 26.0   \\
   (*) + Lasso & 6.7M & 47.6 & 46.6  & 43.1 & 33.2 & 47.9 & 41.5 & 27.5 & 27.0 \\
\midrule
  Full Model Batch+Incr. & 25.9M & \multicolumn{2}{c|}{50.6} &  \multicolumn{2}{c|}{41.8} &  \multicolumn{2}{c|}{52.6} &  \multicolumn{2}{c}{27.0}\\
  (*) + Lasso Batch+Incr. & 9.2M  & \multicolumn{2}{c|}{51.3} &  \multicolumn{2}{c|}{39.1} &  \multicolumn{2}{c|}{51.1} &  \multicolumn{2}{c}{27.6}\\
\midrule
\midrule
  \multicolumn{2}{l|}{\small Repetition Rate Source} & \multicolumn{2}{c|}{{\small 11.0}} & \multicolumn{2}{c|}{{\small 8.8}} & \multicolumn{2}{c|}{{\small 18.3}} & \multicolumn{2}{c}{{\small 9.2}}\\
  \multicolumn{2}{l|}{\small Repetition Rate Target}  & \multicolumn{2}{c|}{{\small 9.4}} & \multicolumn{2}{c|}{{\small 8.7}} & \multicolumn{2}{c|}{{\small 17.5}} & \multicolumn{2}{c}{{\small 7.5}}\\
\bottomrule
\end{tabular}
\caption{Experimental results  in \BLEU (\%) on the English$\to$German data. We evaluate changes to each region of the network separately. In combination with sparse output projection, we also evaluate a \emph{fixed} selection of parameters chosen by thresholding and a set selected dynamically for each data set using group \emph{lasso}. The two bottom rows show repetition rates in \% for the source and target sides of the test data.}
\label{tab:res}
\end{center}
\end{table*}

\subsection{Offset Tensors}

In a personalized translation service, adapted models need to be loaded quickly, so a space-efficient representation is critical for time efficiency as well.
Production speed requirements using contemporary cloud hardware limit model sizes to roughly 10M parameters per user, while a high-quality baseline Transformer model typically requires 35M parameters or more.
We propose to store the parameters of an adapted model as an offset from the baseline model.
Each tensor is a sum $W = W_b + W_u$, where $W_b$ is from the baseline model and is shared across all adapted models, while the offset $W_u$ is specific to an individual user domain. Space efficiency is achieved by only storing $W_u$ for a subset of tensors and setting the rest of the offset tensors to zero.

One approach to achieving model sparsity is to manually partition the network into a small number of regions and systematically evaluate translation performance when storing offsets for only one region.
We define five distinct regions, which are evaluated in isolation: Outer layers (the first and last layers of both encoder and decoder), inner layers (all the remaining layers), the two embedding matrices $X_e$ and $Y_e$, and the output projection matrix $Y_o$.
The latter three are each stored as a single matrix and each contributes 10.3M parameters to the full model size in English$\to$German.
During adaptation, the embedding matrices are only updated for vocabulary present in the training examples, and so the offsets can be stored efficiently as a sparse collection of columns.
The same principle can be applied to the output projection matrix by only updating parameters corresponding to vocabulary items that appears in the adaptation examples (denoted \emph{Sparse Output Proj.} in Table \ref{tab:res}).

A second approach to achieving model sparsity is to use a procedure to select the subset of offset tensors that are stored.
For example, we evaluate a simple policy that stores an offset for all tensors whose average change in parameter values is higher than a threshold.
This set is selected on a development domain and held fixed for all other domains. We refer to this method as \emph{fixed} adaptation.

\subsection{Tensor Selection via Group Lasso}

A group sparse regularization penalty such as group lasso can be applied to the offset tensors for simultaneous regularization and tensor selection.
This penalty drives entire offset tensors to zero, so that they do not need to be stored or loaded.
We add the following regularization term to the loss function \cite{Scardapane:2017:GSR:3067301.3067328}:
\begin{align}
R_{\ell_{1,2}} (\mathcal{T}) &= \sum_{T \in \mathcal{T}} \sqrt{|T|} \|\Delta T\|_2\\
 \|\Delta T\|_2 &= \sum_{\tau \in T} \Delta\tau^2
 \end{align}

 Here, each tensor corresponds to one group. $\mathcal{T}$ denotes the set of all tensors in the model, $\tau \in T$ the set of all weights within a single tensor and $\Delta\tau$ the size of the offset for $\tau$.
 Note that we are regularizing the \emph{difference} between the parameters of the adapted model and the baseline model, rather than regularizing the full network parameters directly.
 In this way, we maintain the expressive power of the full network while minimizing the size of the adapted models. Group lasso regularization is equivalent to $\ell_1$ regularization when the group size is 1. Sparsity among groups is encouraged because the $\ell_1$ norm serves as a convex proxy for the $\ell_0$ norm, which would explicitly penalize the number of non-zero elements \citep{yuan2006model}.
 To facilitate tensor selection, we define a threshold $\vartheta$ to clip offset tensors $\Delta T$ with average weight $\frac{1}{|T|} \sum_{\tau \in T} \Delta\tau < \vartheta$ to zero.
 Both the threshold $\vartheta$ and the regularization weight $\lambda$ were manually tuned on a development domain and set to $\vartheta = 10^{-4}$ and $\lambda = 10^{-6}$. We apply clipping to all tensors except the embedding and output projection matrices $X_e$, $Y_e$ and $Y_o$. As our production constraints allow us to retain only one of the three, we pre-select the sparse output projection as part of the model and exclude the embedding matrices from adaptation. This method will be denoted as \emph{Lasso}.

%% file: experiments.tex
\section{\label{sec:experiments}Experiments}

\subsection{Data}

We first evaluate all techniques on an English$\to$German Transformer network trained on 98M parallel sentence pairs. We apply byte pair encoding \cite{sennrich-haddow-birch:2016:P16-12} separately to each language and obtain vocabularies with 40K unique tokens each. We refer to the unadapted model as \emph{Baseline}.
We evaluate on four domains.
For development, we use a data set labeled \emph{User1} that was gathered from a user of the browser-based CAT (computer-aided translation) tool Lilt\footnote{\scriptsize\url{https://lilt.com}} and contains documents from the financial domain with 48K segments for batch adaptation and 1790 segments for testing and incremental adaptation.
We further evaluate on a second user test set \emph{User2} (technical support, 31k batch adaptation, 1000 test segments); the public Autodesk corpus\footnote{\scriptsize\url{https://autodesk.app.box.com/Autodesk-PostEditing}}, where we select the first 20k segments for batch adaptation and the next 1000 segments for testing; and the IWSLT corpus\footnote{\scriptsize\url{http://workshop2017.iwslt.org/}} (semi-technical talks), where we use all provided 206K sentences for batch adaptation and the \emph{dev2010} set (888 sentences) for testing.
The overall best performing compact adaptation technique, group lasso regularization, is further evaluated on six other language pairs trained using production data sets collected from Lilt's user base: English$\leftrightarrow$French, English$\leftrightarrow$Russian and English$\leftrightarrow$Chinese.
Adaptation is performed on user data from various domains (technical manuals, finance, legal), each with 8k-10k segments for batch adaptation and 2000 segments for testing and incremental adaptation.
Translation quality is evaluated using the cased \BLEU \cite{papineni02:bleu} measure.

\subsection{Results}

\begin{table*}[!tbp]
\begin{center}
%\small
\begin{tabular}{l|cccccc|c}
\toprule
  & en$\rightarrow$fr &  fr$\rightarrow$en &   en$\rightarrow$ru &  ru$\rightarrow$en &  en$\rightarrow$zh &  zh$\rightarrow$en & Avg.\\
\midrule
%baseline & 28.83 & 35.83 & 10.74 & 29.17 & 19.94 & 18.89 & 23.90\\
%full model batch+incr. & 36.63  & 49.62 & 21.02 & 42.09 & 40.61 & 46.63 & 39.43\\
%(*) + lasso batch+incr. & 36.02 & 46.28 & 19.79 & 42.66 & 39.34 & 45.01 & 38.18 \\
Baseline & 28.8 & 35.8 & 10.7 & 29.2 & 19.9 & 18.9 & 23.9\\
Full Model Batch+Incr. & 36.6  & 49.6 & 21.0 & 42.1 & 40.6 & 46.6 & 39.4\\
(*) + Lasso Batch+Incr. & 36.0 & 46.3 & 19.8 & 42.7 & 39.3 & 45.0 & 38.2 \\

\bottomrule
\end{tabular}
\caption{Experimental results in \BLEU (\%) on six production language pairs. We compare the unadapted baseline model with a full model and the model with sparse output projection and group lasso, both with application of batch and incremental adaptation. }
\label{tab:resProd}
\end{center}
\end{table*}

Table~\ref{tab:res} shows English$\to$German results. Full model adaptation, where all offsets are stored, improves over the baseline in all cases to various degrees.
This full model contains only 25.8M parameters, as offsets for both embedding matrices are stored as sparse collections of columns for the vocabulary present in the adaptation data.
Next, we evaluate the impact of storing offsets only for one region at a time.
We observe that among the three vocabulary matrices, the output projection $Y_o$ has the strongest impact on quality, which is not diminished by storing a sparse variant that is restricted only to observed vocabulary.

In addition, we evaluate two methods of choosing a subset of tensors procedurally.
We first experiment with a fixed subset of tensor offsets that was chosen by selecting all tensors for which parameters were offset by more than 0.002 on average after batch adaptation on the \emph{User1} data set.
This simple procedure approaches the performance of full model adaptation, but stores only 27\% of its parameters.
Dynamically selecting tensor offsets for each data set using group lasso regularization improves performance on 6 out of 8 data conditions.

The combination of batch and incremental adaptation yields further improvements, with the exception of the \emph{User2} and \emph{IWSLT} tasks, where incremental adaptation overall performs not as well as batch adaptation. For these tasks, both tests sets exhibit lower repetition
rates\footnote{Repetition rates have been confirmed to be a suitable indicator for gains through incremental adaptation in numerous works \citep{wuebker15:incrementalAdaptation,bertoldi:2014:onlineAdaptation}.} \citep{cettolo14:repetitionRate} than the test sets for the two other tasks (see the two bottom lines in Table \ref{tab:res}). The \emph{User2} test set is furthermore a random sample of non-consecutive text from a translation memory, which is suboptimal for incremental learning.

Altogether, we are able to achieve translation performance similar to full model adaptation with 25\% of the total network parameters.
Note that due to the selection of entire tensors with groupwise regularization, there is nearly zero space overhead incurred by storing a sparse set of offset tensors.

Table \ref{tab:resProd} confirms our main findings on six other language pairs. We observe average improvements of 14.3 \BLEU with our final compact model, which compares to 15.5 \BLEU for full model adaptation.

%% file: 9-conclusion.tex
\section{Conclusion}

We describe an efficient approach to personalized machine translation that stores a sparse set of tensor offsets for each user domain.
Group lasso regularization applied to the offsets during adaptation achieves high space and time efficiency while yielding translation performance close to a full adapted model, for both batch and incremental adaptation and their combination.

%\section*{Acknowledgments}